\documentclass[10pt,twocolumn,letterpaper]{article}\usepackage[pagenumbers]{cvpr}

\usepackage[T1]{fontenc}
\usepackage{times,microtype,amsmath,amssymb,booktabs,array,graphicx,enumitem,placeins}
\usepackage[numbers,sort&compress]{natbib}
\usepackage[colorlinks=true,linkcolor=blue,citecolor=blue,urlcolor=blue]{hyperref}

\title{DynaCue: Comparing Motion and Timing Between Groups of Videos}
\author{Hao Chen\\Department of Statistics\\University of California, Davis}\date{}
\hypersetup{pdfauthor={Hao Chen},pdftitle={DynaCue: Comparing Motion and Timing Between Groups of Videos}}

\begin{document}
\maketitle
\begin{abstract}
A score that distinguishes videos with different temporal-consistency ratings can miss differences associated with motion preference. VideoScore's temporal consistency score distinguished higher- from lower-rated Pika videos in 95\% of comparisons with 15 videos per group, yet could not detect the difference between preferred videos and their same-prompt alternatives in any of the 200 comparisons of 41 VideoGen-RewardBench pairs. We introduce DynaCue, a two-sample permutation test that combines complementary measurements of motion, frame continuity, and appearance to compare groups of videos without choosing a single cue in advance. DynaCue measures local changes in optical flow both before and after accounting for each clip's usual activity, captures interruptions through frame differences and freeze counts, and incorporates scores for motion and appearance. In controlled experiments, DynaCue detected group differences across five interventions affecting playback continuity, displacement, or rhythm. In real-world evaluations of generated videos where viewers rated temporal consistency or chose which of two videos they preferred for its motion, DynaCue detected group differences in at least 80\% of resampled comparisons on each of the three datasets at the smaller sample sizes studied. Every other method evaluated on all three had a detection rate of 60\% or less on at least one dataset.
\end{abstract}
\begin{figure*}[t]\centering
\includegraphics[width=\textwidth]{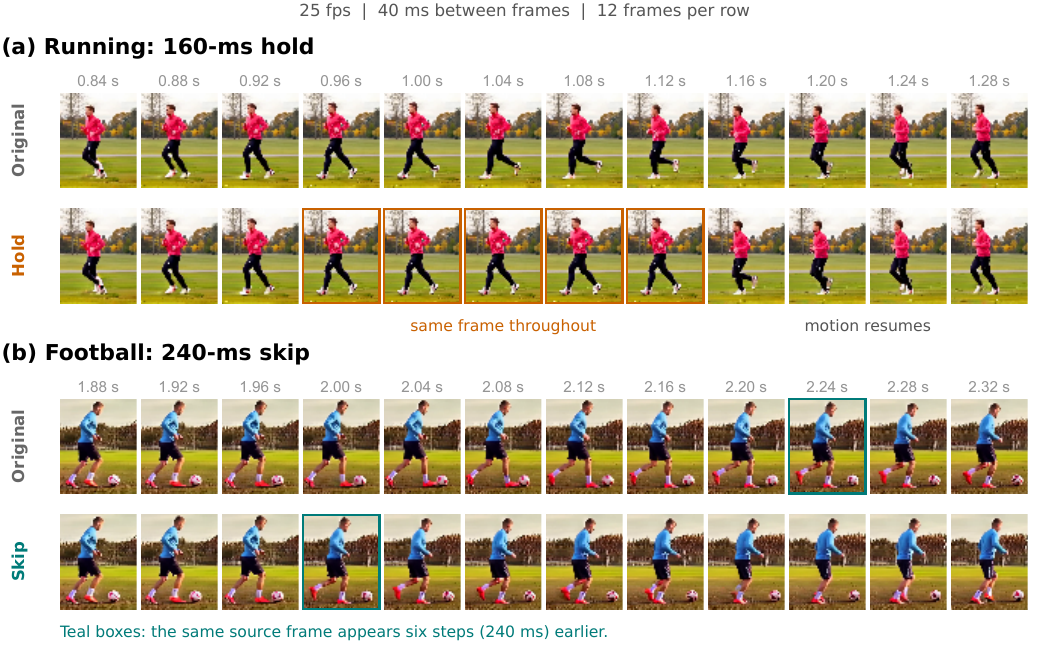}
\caption{A frame hold and a forward skip in StepVideo clips from VBench-2.0~\citep{vbench2release,vbench2paper}. Each column shows the same crop and output time in the original and edited clips. Successive columns are 40 ms apart at the native 25 fps. (a) The edit holds the runner on source frame 24 from 0.96 to 1.12 s (orange). (b) The football player reaches source frame 56 at 2.00 s in the edited clip and 2.24 s in the original clip (teal). At 8 Hz the original football clip uses source frames 50 and 53 at 2.000 and 2.125 s, while the edited clip uses frames 56 and 59.}\label{fig:fixedteaser}
\end{figure*}
\section{Introduction}
Evaluating a video generator adds a question that does not arise for still images: does the action unfold plausibly over time? Every frame can look convincing on its own, even when the action pauses or jumps ahead. Figure~\ref{fig:fixedteaser} illustrates both interruptions. The runner hangs in mid-stride as the same frames repeat, while the football player jumps ahead because part of his stride is missing. Either interruption can be easy to miss in playback. Recognizing it requires following the movement across successive frames.

Even when the movement in each clip looks plausible, a generator can miss some of the ways an action can unfold. For example, Denton and Fergus~\citep{dentonfergus} asked whether a video predictor captured the range of directions a digit could take after hitting a wall. They compared the distribution of post-impact directions in predicted trajectories with that in simulated trajectories.

For playback continuity, both displacement and frame repetition can be informative. In Fig.~\ref{fig:fixedteaser}, a large displacement between successive frames can occur both when the football player's stride skips ahead and when the runner catches up after the pause. Repeated frames provide an additional cue to the pause. Comparing samples of videos with higher and lower temporal-consistency ratings lets us ask whether interruptions are more frequent or more pronounced in one group. The ratings alone leave open which aspect of the dynamics will distinguish the groups. Our goal is to build a test that can detect a range of differences in motion and temporal consistency between the two groups.

We introduce DynaCue, a two-sample test that uses several complementary measurements, including local changes in optical flow, differences between adjacent frames, freeze counts, and VBench scores for motion and appearance~\citep{vbench}. In controlled experiments, DynaCue detected changes in playback continuity, displacement, and rhythm. In three real-world evaluations of generated videos, it also detected differences between groups formed from viewers' temporal-consistency ratings or motion preferences~\citep{videoscore,evalcrafterrelease,videorewardbench}.

\paragraph{Related work.}
Collection metrics compare two groups through their video features. Fr\'echet Video Distance (I3D-FVD) compares the means and covariances of features from an inflated 3D convolutional network~\citep{unterthiner}. Content-debiased FVD (CD-FVD) uses features from a video masked autoencoder (VideoMAE) to reduce sensitivity to visual content~\citep{ge}. Fr\'echet Video Motion Distance (FVMD) uses velocity and acceleration histograms from tracked points~\citep{liu}. JEPA Embedding Distance (JEDi) applies a polynomial kernel to embeddings from a video joint-embedding predictive architecture (V-JEPA)~\citep{jedi}.

Per-video scores describe particular aspects of a clip. EvalCrafter combines frame similarity, warping error and face consistency into a temporal-consistency score~\citep{evalcrafterrelease}. VBench reports separate measures of appearance and motion~\citep{vbench}, while VideoScore and VideoScore2 predict multiple dimensions of video quality~\citep{videoscore,videoscore2}. Seeing Fast and Slow detects local changes in playback speed~\citep{seeingfastslow}.

DynaCue draws on two-sample tests that compare similarities within groups. The rank in similarity graph edge-count test (RISE) ranks edges in a similarity graph~\citep{zhou}, while the generalized permutation-based kernel test (GPK) measures similarity with a kernel~\citep{song}. The Z-resolved Integrated Diagnostic (ZID) combines graph and kernel contrasts for a single representation at multiple bandwidths~\citep{zid}. Bandwidth matters because the sensitivity of a kernel test can depend on the scale of the difference and the available sample size~\citep{finitekernel}.

\section{The DynaCue test}
DynaCue compares the two groups using motion, continuity, and appearance measurements in two branches, then computes a single $p$-value by permuting whole-video labels as the study design allows.

\paragraph{Task and experimental unit.}
Each panel consists of two samples of videos. In controlled experiments, we draw disjoint samples from the same pool, leave one group unedited, and independently edit each clip in the other with probability $\rho$. In VideoFeedback and Pika, we compare higher-rated with lower-rated clips. In RewardBench, each prompt contributes one motion-preferred clip to one group and its paired alternative to the other.

Under the two-sample null, the video measurements have the same joint distribution in the two groups. We always reassign whole videos, keeping all frames, windows and scores of a video together. The study design determines which reassignments are allowed. In VideoFeedback and Pika, the null requires exchangeability of rating labels within matching strata. In RewardBench, we condition on the selected prompt pairs and on the preference counts for each generator pairing. The observed assignment must be exchangeable with the other assignments that satisfy those restrictions.

\paragraph{Observation and representation.}
Our motion representation, burst12, summarizes local variation in optical flow. In the controlled experiments we observe four seconds at 8 Hz, including both endpoints, and select the nearest native frame at each of the 33 observation times. We resize each $256\times256$ RGB frame to a $128\times128$ grayscale image for Farneb\"ack optical flow~\citep{farneback}. We divide displacements by image width and elapsed time. For each transition we record the median, 90th and 99th percentiles of speed, mean horizontal and vertical flow, and mean residual speed after subtracting the spatial median flow. Within each eight-frame window, two quantities describe each of these six signals: the largest absolute change between consecutive transitions and the difference between its 90th percentile and median. These six pairs of summaries form a 12-dimensional window descriptor $g_w$. Windows advance by four frames and include the final available window. For each coordinate we measure how far its most unusual window departs from its typical window:
\begin{equation}
 b_j(X)=\max_w|g_{wj}-\operatorname{median}_u g_{uj}|,\quad j=1,\ldots,12.
\end{equation}
The resulting 12-dimensional vector is \emph{burst12}. It preserves the scale of local motion variation in units of image widths per second. An activity-relative version, $b_{\rm rel}(X)=b(X)/a(X)$, measures the same excursions relative to the clip's usual speed. We compute $a(X)$ by averaging the spatial median speed across transitions in each window and then averaging across windows. This normalization requires finite, positive activity in every video in the panel. If any relative vector is undefined, the full test yields no $p$-value. To compare burst vectors, we scale each coordinate by its standard deviation across the pooled videos, using population normalization and a scale of one for standard deviations at most $10^{-10}$, and apply weight $1/\sqrt{12}$. We order $g_w$ by listing the six maximum-change coordinates first, then the six upper-tail differences. Within each block we use the transition-signal order above.

Motion confined to part of the image can raise the 99th percentile of pixel speeds while the median stays near zero. A camera pan affects much of the image and can also register in the spatial median. The maximum in $b_j$ retains the size of a departure wherever it occurs in the clip. Fast, steady motion can yield a small burst12 vector when all the windows have similar summaries.

\paragraph{Measurements in the two branches.}
The first branch compares the two groups' burst12 vectors using four of the six contrasts in standard ZID. We started with ZID because it showed sensitivity to various types of distributional change in prior work on generative evaluation~\citep{zid}. Standard ZID constructs a RISE graph and two Gaussian kernels on the pooled vectors. The kernel bandwidths are the median positive distance between pooled vectors and 0.175 times that median. For each similarity measure, ZID sums the similarities within each group and computes two standardized contrasts from the centered sums. The contrast $W$ uses their weighted sum and $D$ their difference.

All six contrasts use the same burst12 vectors, which describe the magnitudes of local motion changes. A forward skip may produce only a modest displacement in a slow-moving clip, even when that displacement is large relative to the clip's usual motion. A pause leaves another cue that burst12 does not record directly: successive frames may repeat. We therefore added a second branch that measures motion relative to each clip's usual activity, captures interruptions through adjacent-frame differences and freeze counts, and compares six VBench scores for appearance and motion~\citep{vbench}. The six scores form one vector per video and cover subject consistency, background consistency, motion smoothness, dynamic degree, aesthetic quality and imaging quality.

For each clip, MAD is the smallest mean absolute RGB difference between adjacent frames. For every allowed allocation of videos, we take the absolute difference between the two group means of MAD and rank the observed difference against those values. FF counts freeze episodes reported by FFmpeg's \texttt{freezedetect}~\citep{ffmpegfilters,ffmpegsource}. We compare the distributions of these counts using the scalar energy statistic~\citep{energy}.

The initial test had six terms in each branch. The first used all six ZID contrasts on burst12. The second used the three $W$ contrasts on activity-relative burst12, together with one term each for the VBench score vector, MAD and FF. The ablation study in Sec.~\ref{sec:selection} led to the final four terms per branch. The first retains graph $W$, both $W$ and $D$ for the median-bandwidth kernel, and $D$ for the small-bandwidth kernel. The second retains small-bandwidth kernel $W$ on activity-relative burst12, median-bandwidth kernel $W$ on the VBench scores, MAD and FF.

\paragraph{One decision across both branches.}
For each label allocation $b$, DynaCue combines the two-sided normal tail areas of the four standardized contrasts in its first branch with Simes' rule~\citep{simes}. For four ordered tail areas $q_{(1)}\le\cdots\le q_{(4)}$, the Simes score is $\min_{1\le j\le4}4q_{(j)}/j$. The second branch first ranks each constituent statistic across those same allocations and applies Simes' rule to the four rank probabilities. For each branch we then count the fraction of allocations with a Simes score at least as small as the one for $b$. Denote these two fractions by $a_b$ and $c_b$. For each allocation we compute the harmonic mean $p$-value (HMP)~\citep{wilsonhmp}, $h_b=2a_bc_b/(a_b+c_b)$. Smaller values give stronger evidence. Of the $B$ label allocations, $b=0$ denotes the observed assignment. Its final permutation $p$-value is
\[
 \widehat p=B^{-1}\sum_{b=0}^{B-1}\mathbf{1}\{h_b\le h_0\}.
\]
All measurements from a video move together under relabeling, preserving their dependence. The count includes the observed allocation and ties~\citep{phipson,randomperm}. We use $B=500$ for the controlled experiments, 4,000 for VideoFeedback and Pika, and 5,000 for RewardBench. We report the percentage of panels with $\widehat p\le .05$: the detection rate under an edit or a human-label contrast, and the rejection rate under randomized labels.

\paragraph{External comparisons.}
We evaluate each external method on the same panels and label allocations as DynaCue. We test each of the six VBench dimensions, five VideoScore dimensions, and three VideoScore2 dimensions separately. EvalCrafter contributes its published temporal-consistency score, which combines frame similarity, warping error, and face consistency. For Seeing Fast and Slow, we use the peak speed-change probability across windows on the shared 8-Hz input. For each of these per-video scores, the test statistic is the absolute difference between the two group means.

\begin{figure*}[t]\centering
\includegraphics[width=\textwidth]{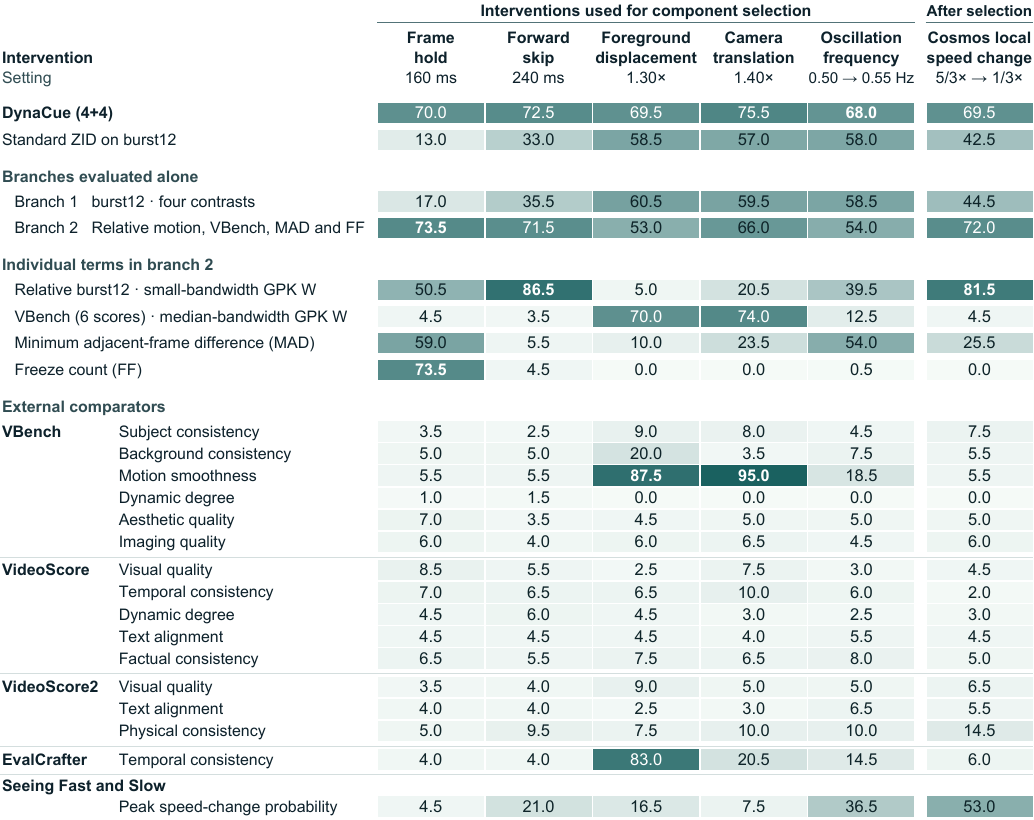}
\caption{Detection of controlled changes in playback and motion. Each cell gives the percentage of 200 panels with $p\le .05$ on the same video draws for all methods. The first five columns show one setting per intervention, chosen for intermediate detection by the initial twelve-term test. The displacement and translation settings are multiples of the original movement amplitude. The last column evaluates a local speed change on Cosmos clips~\citep{videophy2} after component selection. Playback runs at $5/3$ speed for half a second, then at $1/3$ speed for half a second. Each group contains 20 clips for holds and 40 otherwise. The edit probability is .50 for frequency and .75 otherwise. Standard ZID on burst12 shows how often the six-contrast test detects each intervention using local optical-flow variation alone. Branch 1 uses the four contrasts retained in DynaCue on the same features. The rows below the two branches show the four distinct measurements in the second branch. Bold marks the highest rate in each column.}\label{fig:controlled}
\end{figure*}
\section{Controlled interventions and component selection}\label{sec:selection}
\paragraph{Five interventions in playback continuity, displacement, and rhythm.}
Figure~\ref{fig:controlled} compares how often DynaCue, its components, and external scores detect each intervention. We edit 100 StepVideo clips from the VBench-2.0 video release~\citep{vbench2release,vbench2paper} by holding the source clock on one frame or advancing it by a fixed interval. Across successive clips, both edits start at 1, 2 or 3 seconds in rotation. Each onset coincides with an observation on the 8-Hz grid. After a hold, playback jumps ahead to rejoin the original clock. In three single-factor synthetic experiments we vary foreground displacement, whole-image camera excursion, or the frequency of a subject moving back and forth. Within each synthetic comparison the scene and texture are fixed by source. In the frequency experiment the object travels the same distance on each stroke. Changing the period also changes its speed. Each signal condition uses 200 panels, with identical source draws and label allocations across methods.

\paragraph{Selection across interventions.}
The five interventions contributed 68 conditions across intervention strengths, sample sizes and edit probabilities. We started with standard six-contrast ZID on burst12, using $W$ and $D$ for the RISE graph and each of the two Gaussian kernels. Exploring the five interventions suggested three ways to complement burst magnitude. We used the three $W$ contrasts on activity-relative burst12 to capture motion unusual for a given clip, MAD and freeze count to register interruptions, and one joint VBench term to capture changes in appearance and motion. At each step we compared every single deletion with its parent on the same panels and label allocations.

To select terms, we ranked deletions by mean detection across five representative settings, one per intervention. At each setting the initial test detected close to 70\% of panels, leaving room to observe gains and losses. To summarize performance across the full range of strengths, sample sizes, and edit probabilities, we also report a mean over all 68 conditions. We first average within each intervention and then average the five means, so the four hold conditions receive the same total weight as the 28 foreground-displacement conditions.

Mean detection on the five settings rose from 68.2\% with twelve terms to 71.1\% with nine. Removing graph $D$ from the first branch preserved that mean with eight terms and raised the 68-condition mean from 54.02\% to 54.25\%. We stopped at four terms per branch because every further single deletion reduced the five-setting mean. The best seven-term variant removed small-bandwidth kernel $D$ from the first branch. Its mean was 70.8\%, with foreground-displacement detection falling from 69.5\% to 67.0\%. Its 68-condition mean was slightly higher (54.41\%). The retained measurements cover different changes: removing the activity-relative motion term reduced skip detection by 43 percentage points, while removing the burst12 branch reduced detection of foreground displacement, camera translation, and oscillation frequency (Table~\ref{tab:deletions}).

\begin{table}[!htbp]\centering
\caption{DynaCue detection with and without individual terms or branches. The first row gives detection rates (\%) for the full test. For each deletion, the five intervention columns give changes from the full test in percentage points. Mean gives the detection rate (\%) averaged equally across the five settings in every row. All rows use the same 200 panels and label allocations per setting as in Fig.~\ref{fig:controlled}.}\label{tab:deletions}
\footnotesize\setlength{\tabcolsep}{0pt}
\begin{tabular*}{\columnwidth}{@{\extracolsep{\fill}}lrrrrrr@{}}\toprule
Removed & Hold & Skip & Displ. & Camera & Freq. & Mean\\\midrule
None (full $4+4$) & 70.0 & 72.5 & 69.5 & 75.5 & 68.0 & \textbf{71.1}\\\midrule
\multicolumn{7}{@{}l}{\itshape First-branch term (burst12)}\\
Graph $W$ & $-1.5$ & $-3.5$ & $+1.0$ & $+2.5$ & $-6.5$ & 69.5\\
Median kernel $W$ & $+1.0$ & $+0.5$ & 0.0 & $-3.5$ & $-1.5$ & 70.4\\
Median kernel $D$ & 0.0 & $+0.5$ & $-3.0$ & $-1.5$ & $+1.0$ & 70.5\\
Small kernel $D$ & $+0.5$ & 0.0 & $-2.5$ & 0.0 & $+0.5$ & 70.8\\
\addlinespace[2pt]
\multicolumn{7}{@{}l}{\itshape Second-branch term}\\
Relative, small $W$ & $-1.5$ & $-43.0$ & $+2.5$ & $+1.5$ & $-2.0$ & 62.6\\
VBench, median $W$ & $+3.0$ & $+2.0$ & $-19.0$ & $-19.5$ & $+3.0$ & 65.0\\
MAD & $-5.5$ & $+2.0$ & $+2.0$ & $+2.5$ & $-9.0$ & 69.5\\
Freeze count & $-11.5$ & $+1.5$ & 0.0 & 0.0 & $+0.5$ & 69.2\\
\addlinespace[2pt]
\multicolumn{7}{@{}l}{\itshape Whole branch}\\
First branch & $+3.5$ & $-1.0$ & $-16.5$ & $-9.5$ & $-14.0$ & 63.6\\
Second branch & $-53.0$ & $-37.0$ & $-9.0$ & $-16.0$ & $-9.5$ & 46.2\\
\bottomrule\end{tabular*}\end{table}
\paragraph{Contributions of individual terms and branches.}
Which terms helped depended on what changed in the video (Table~\ref{tab:deletions}). Removing graph $W$ from the burst12 branch caused its largest loss on oscillation frequency. Removing the VBench term from the second branch reduced detection by 19 percentage points for foreground displacement and 19.5 for camera translation, while increasing hold detection by 3 points.

Table~\ref{tab:branch-comparison} compares the two-branch test with two ways of combining all eight terms in a single Simes step. Flat Simes A uses the same inputs as DynaCue: four two-sided normal-tail probabilities from the first branch and four permutation probabilities from the second. Flat Simes B uses permutation probabilities for all eight terms. All three tests use the same label allocations. Each computes its final $p$-value as the fraction of allocations with a combined score at least as small as its observed score. With identical inputs, combining the terms in two branches increased detection by 2.0 to 4.0 percentage points across the five settings.

\begin{table}[t]\centering
\caption{Detection rates (\%) under three rules for combining the eight terms. Each comparison uses the same 200 panels per setting as Fig.~\ref{fig:controlled}. Mean weights the five settings equally.}
\label{tab:branch-comparison}
\small\setlength{\tabcolsep}{1.45pt}
\begin{tabular*}{\columnwidth}{@{\extracolsep{\fill}}lccc@{}}\toprule
 & & \multicolumn{2}{c}{Flat Simes}\\\cmidrule(lr){3-4} Intervention & DynaCue & A & B\\\midrule
Hold & 70.0 & 68.0 & 69.5\\
Forward skip & 72.5 & 70.0 & 71.0\\
Foreground displacement & 69.5 & 66.0 & 66.0\\
Camera translation & 75.5 & 73.5 & 70.5\\
Oscillation frequency & 68.0 & 64.0 & 61.0\\
\midrule Mean & 71.1 & 68.3 & 67.6\\\bottomrule\end{tabular*}\end{table}

\paragraph{A hold and a skip leave different evidence.}
Holding a frame creates a repetition and a catch-up displacement when playback resumes. At the 8-Hz observation rate, which repetitions remain visible depends on when the hold begins (Fig.~\ref{fig:fixedteaser}). With 20 clips per group and $\rho=.75$, the two interruption measures together detected a 160-ms hold in 75.0\% of panels, compared with 50.5\% for relative motion alone and 70.0\% for DynaCue.

A 240-ms forward skip produced the opposite pattern. At $n=40$ and $\rho=.75$, relative motion alone detected 86.5\% of panels and the two interruption measures detected 6.5\%. Removing the relative-motion term from DynaCue reduced detection from 72.5\% to 29.5\%. With the full test, lowering the edit probability in the intervention group from .75 to .25 reduced detection of the same skip from 72.5\% to 14.5\% (Table~\ref{tab:rho}).

VBench motion smoothness detected 87.5\% of foreground displacement panels and 95.0\% of camera translation panels, but only 5.5\% for each of the two playback interruptions (Fig.~\ref{fig:controlled}). Across the five settings used for selection, DynaCue's detection rates ranged from 68.0\% to 75.5\%.

\begin{table}[!htbp]\centering
\caption{Detection of a 160-ms hold or a 240-ms forward skip as edits become more common. Entries give the percentage of 200 panels with $p\le .05$, with $n$ StepVideo clips per group and probability $\rho$ of editing each clip in the intervention group. The null row gives rejection rates on 1,000 unedited panels per group size, shared by the two edits.}\label{tab:rho}
\small\setlength{\tabcolsep}{4pt}
\begin{tabular}{@{}lrrrr@{}}\toprule
& \multicolumn{2}{c}{Hold (160 ms)} & \multicolumn{2}{c}{Forward skip (240 ms)}\\
\cmidrule(lr){2-3}\cmidrule(l){4-5}
Edit probability & $n=20$ & $n=40$ & $n=20$ & $n=40$\\\midrule
Null & 5.8 & 5.0 & 5.8 & 5.0\\
\addlinespace[2pt]
.125 & 7.5 & 6.5 & 6.0 & 6.5\\
.25 & 13.5 & 16.5 & 7.5 & 14.5\\
.375 & 19.0 & 35.0 & 9.0 & 20.5\\
.50 & 30.5 & 61.5 & 14.5 & 31.5\\
.625 & 49.0 & 81.5 & 22.5 & 51.0\\
.75 & 70.0 & 97.5 & 33.5 & 72.5\\
.875 & 87.0 & 99.5 & 48.0 & 93.5\\
1.00 & 100.0 & 100.0 & 71.5 & 100.0\\
\bottomrule\end{tabular}\end{table}

\paragraph{Local playback speed after component selection.}
We then evaluated the eight-term test on Cosmos clips with a brief acceleration followed by a slowdown (Fig.~\ref{fig:controlled}, last column). With 40 clips per group and edit probability .75, DynaCue detected 18.0\% of panels when playback ran at $4/3$ then $2/3$ of its original rate. Increasing the contrast to $5/3$ then $1/3$ raised detection to 69.5\%. On unedited clips, the test rejected in 45 of 1,000 panels. At the rates shown in Fig.~\ref{fig:controlled}, the burst12 branch alone detected 44.5\% and the second branch alone detected 72.0\%.

\begin{figure}[!htbp]\centering
\includegraphics[width=\columnwidth]{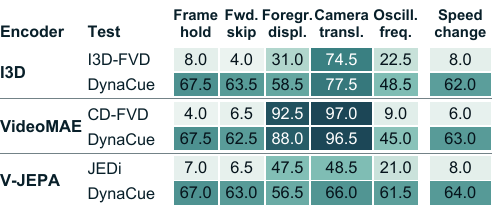}
\caption{Replacing the first-branch motion features with pretrained video embeddings. Within each encoder pair, the upper row gives the published metric and the lower row gives the full DynaCue test using the same encoder output in its first branch. The second branch is fixed. Values are detection rates (\%) on the same 200 panels per column as in Fig.~\ref{fig:controlled}, with whole videos permuted together with their windows.}\label{fig:native-encoders}
\end{figure}

\paragraph{Pretrained video encoders in the first branch.}
Figure~\ref{fig:native-encoders} pairs each published metric with a version of DynaCue that uses the same encoder output in its first branch. We apply the four contrasts selected using burst12 to the new encoder output, keeping the second branch and label allocations fixed. I3D-FVD~\citep{unterthiner} supplies one I3D embedding per clip~\citep{i3d}. For CD-FVD~\citep{ge}, each clip contributes three VideoMAE V2 window embeddings~\citep{videomaev2} to the group distribution. All three move together when we relabel the clip. JEDi~\citep{jedi} supplies one V-JEPA embedding per clip~\citep{vjepa}.

\begin{figure*}[t]\centering
\includegraphics[width=\textwidth]{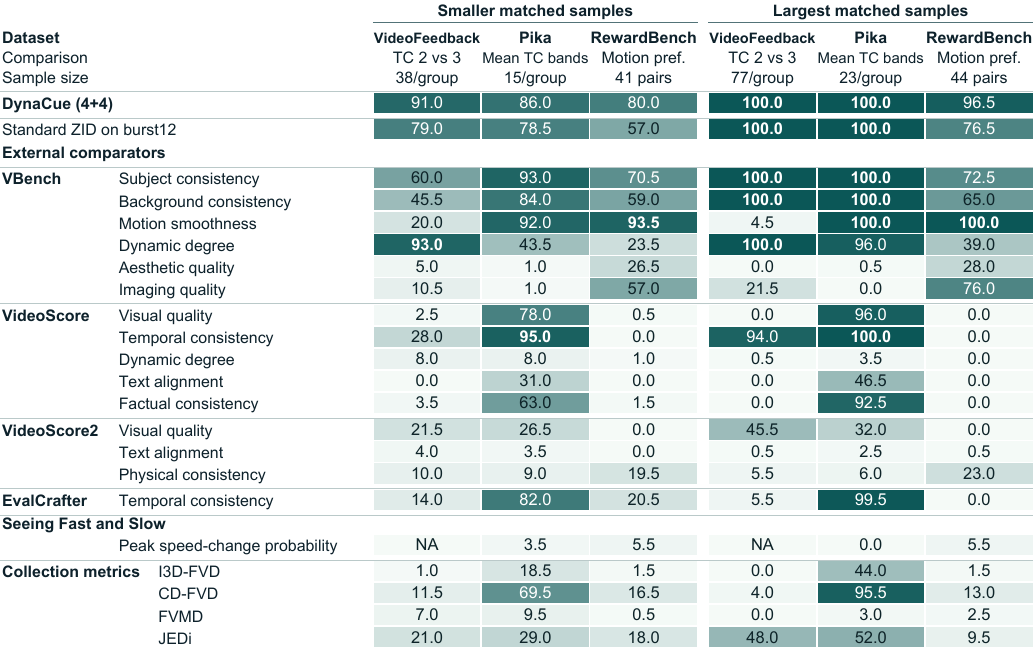}
\caption{Detection of group differences in three human evaluations of generated videos. Standard ZID on burst12 provides a baseline for DynaCue using only local optical-flow variation. The left block uses the smaller matched samples and the right block the largest available within each study pool. VideoFeedback compares temporal-consistency ratings of 2 and 3. Pika compares mean rating bands $[1.8,2.334]$ and $[2.8,3.334]$. RewardBench compares the preferred videos with their same-prompt alternatives for motion. Sample sizes count clips per group for the rating comparisons and prompt pairs for RewardBench. Each cell gives the percentage of 200 panels with $p\le .05$. Bold marks column maxima. NA indicates that the VideoFeedback observations are too short for the speed detector's 2.13-second window.}\label{fig:real}
\end{figure*}

Across the five settings used for component selection, DynaCue averaged 63.1\% detection with I3D, 71.9\% with VideoMAE and 62.8\% with V-JEPA, compared with 71.1\% using burst12. With VideoMAE, CD-FVD detected 4.0\% of holds and 92.5\% of foreground displacement panels. DynaCue, which also uses its second branch, detected 67.5\% and 88.0\% on the same panels.

\section{Temporal-consistency ratings and motion preferences}
We selected the eight terms using only the controlled interventions. Earlier versions of the test had been evaluated on some of the human-rated pools. We applied the selected test to three comparisons defined by human judgments (Fig.~\ref{fig:real}). In VideoFeedback~\citep{videoscore}, we compare clips rated 2 and 3 for temporal consistency, matching on generator and on human ratings of visual quality and dynamic degree. Viewers rated the full video; the test sees its first 16 observed frames (1.875 seconds from first to last). For Pika~\citep{evalcrafterrelease}, the ratings are averages across viewers. We observe native frames $0,3,\ldots,69$ of the 24-fps clips. We compare the bands $[1.8,2.334]$ and $[2.8,3.334]$ within strata defined by rounded visual-quality and motion-quality scores.

VideoGen-RewardBench~\citep{videorewardbench} pairs videos made for the same prompt. We use pairs that tie on visual quality and text alignment but differ in motion preference. We observe each clip on an 8-Hz grid up to its last native frame. To sample a panel, we assign independent uniform random costs to the eligible pairs and select the minimum-cost set satisfying the preference quotas and using each prompt at most once. At 44 pairs the two preference directions balance within every generator pairing. At 41 pairs one pairing has one extra preference in one direction. The pairing and direction rotate across draws; permutations preserve the actual counts in each draw. For I3D-FVD, CD-FVD, FVMD, and JEDi, we recompute the published collection statistic under each label allocation.

\paragraph{Detection across judgments.}
We chose the smaller sample sizes so that the highest observed detection rate among individual scores was near 95\% on each dataset. A different score reached that level in each comparison: VBench dynamic degree for VideoFeedback, VideoScore temporal consistency for Pika, and VBench motion smoothness for RewardBench. Using the same measurements throughout, DynaCue detected differences in 91.0\% of VideoFeedback panels (38 clips per group), 86.0\% of Pika panels (15 per group), and 80.0\% of RewardBench panels (41 pairs). Among the other methods evaluated on all three datasets, VBench subject consistency had the highest minimum rate, 60.0\%.

At the largest matched sample sizes, several methods detected every panel in one or both rating comparisons. DynaCue detected all 200 panels on both VideoFeedback (77 per group) and Pika (23 per group), and 193 of 200 on RewardBench (44 pairs). These rates describe repeated samples from the video pools used for matching. For example, every maximum-size VideoFeedback panel includes all 77 lower-rated clips available for matching, while the sample of higher-rated clips can vary. On RewardBench, 29 pairs recur in all 200 maximum-size panels. At the smaller sizes for VideoFeedback and Pika, we draw stratum slots from the lower rating or score band without replacement, then sample clips without replacement within each group and stratum to match those counts.

\section{Discussion}
DynaCue detected changes in continuity, displacement, and rhythm using the same set of measurements. Repeated frames made freeze counts especially sensitive to holds, while a forward skip produced a sudden displacement that was more readily detected by activity-relative motion. For changes in how far or how often a subject moved, combining the two branches improved detection over either branch alone (Fig.~\ref{fig:controlled}). A targeted measurement can still have the advantage when the edit is known in advance.

In our controlled experiments, frame holds and forward skips began on the 8-Hz sampling grid. A hold of the same duration starting between sampling times can produce a different sequence of observed frames. Detection also depends on how many clips are sampled per group and what fraction contain the intervention (Table~\ref{tab:rho}). In the human comparisons, a rejection gives evidence of a group difference in the measured properties.

False-positive control depends on exchangeability under the study design~\citep{randomperm}. At the nominal 5\% level, DynaCue rejected 4.0--5.8\% of unedited panels across the five controlled interventions and 4.7--6.2\% across six randomized-label controls on the human-rated pools. The latter controls use random labels drawn under the matching restrictions; the rates in Fig.~\ref{fig:real} use the recorded judgments. 

\FloatBarrier
\section*{LLM Usage Statement}
We used OpenAI Codex with GPT-family models to search the literature, develop and debug software, run experiments, check numerical results and their sources, prepare figures, and draft and revise the manuscript. We used two multimodal language models, VideoScore and VideoScore2, to score individual videos in the experiments reported in Figs.~\ref{fig:controlled} and~\ref{fig:real}. The author is responsible for the study design, results and interpretation.


{\small
\begin{thebibliography}{99}

\bibitem{unterthiner} T. Unterthiner, S. van Steenkiste, K. Kurach, R. Marinier, M. Michalski and S. Gelly. Towards accurate generative models of video: A new metric \& challenges. \emph{arXiv:1812.01717}, 2018, revised 2019. \url{https://arxiv.org/abs/1812.01717}.

\bibitem{ge} S. Ge, A. Mahapatra, G. Parmar, J.-Y. Zhu and J.-B. Huang. On the content bias in Fr\'echet video distance. \emph{CVPR}, 2024. \url{https://arxiv.org/abs/2404.12391}.

\bibitem{liu} J. Liu, Y. Qu, Q. Yan, X. Zeng, L. Wang and R. Liao. Fr\'echet video motion distance: A metric for evaluating motion consistency in videos. \emph{arXiv:2407.16124}, 2024. \url{https://arxiv.org/abs/2407.16124}.

\bibitem{jedi} G. Y. Luo, G. M. Favero, Z. H. Luo, A. Jolicoeur-Martineau and C. Pal. Beyond FVD: An enhanced metric for evaluating video generation distribution quality. \emph{ICLR}, 2025. \href{https://proceedings.iclr.cc/paper_files/paper/2025/file/a57483b394a3654f4317051e4ce3b2b8-Paper-Conference.pdf}{Conference paper}.

\bibitem{vbench} Z. Huang et al. VBench: Comprehensive benchmark suite for video generative models. \emph{CVPR}, pp. 21807--21818, 2024. \url{https://github.com/Vchitect/VBench}.

\bibitem{vbench2release} Vchitect. VBench-2.0 sampled videos. Public generated-video dataset, pinned revision \texttt{f9ee15ef\allowbreak cfb86d96\allowbreak bd5379e8\allowbreak d85f8d69\allowbreak b901aad1}. \url{https://huggingface.co/datasets/Vchitect/VBench-2.0_sampled_videos}.

\bibitem{zhou} D. Zhou and H. Chen. A new ranking scheme for modern data and its application to two-sample hypothesis testing. \emph{COLT}, PMLR 195:3615--3668, 2023. \url{https://proceedings.mlr.press/v195/zhou23a.html}.

\bibitem{song} H. Song and H. Chen. Generalized kernel two-sample tests. \emph{Biometrika} 111(3):755--770, 2024. doi:10.1093/biomet/asad068.

\bibitem{zid} H. Chen. \emph{What FID Hides: Detecting, Ranking, and Diagnosing Deviations in Generative Evaluation}. arXiv:2608.24881, 2026.

\bibitem{wilsonhmp} D. J. Wilson. The harmonic mean $p$-value for combining dependent tests. \emph{PNAS} 116(4):1195--1200, 2019. doi:10.1073/pnas.1814092116. Correction: \emph{PNAS} 116(43):21948, 2019, doi:10.1073/pnas.1914128116.

\bibitem{simes} R. J. Simes. An improved Bonferroni procedure for multiple tests of significance. \emph{Biometrika} 73(3):751--754, 1986.

\bibitem{phipson} B. Phipson and G. K. Smyth. Permutation $p$-values should never be zero: Calculating exact $p$-values when permutations are randomly drawn. \emph{Statistical Applications in Genetics and Molecular Biology} 9(1):Article 39, 2010.

\bibitem{farneback} G. Farneb\"ack. Two-frame motion estimation based on polynomial expansion. \emph{SCIA}, LNCS 2749:363--370, 2003. doi:10.1007/3-540-45103-X\_50.

\bibitem{videophy2} H. Bansal, C. Peng, Y. Bitton, R. Goldenberg, A. Grover and K.-W. Chang. VideoPhy-2: A challenging action-centric physical commonsense evaluation in video generation. \emph{ICLR}, 2026. \href{https://proceedings.iclr.cc/paper_files/paper/2026/hash/c02f6a1d5c55e16db50d339dad905b4d-Abstract-Conference.html}{Conference paper}.

\bibitem{videoscore} X. He, D. Jiang, G. Zhang et al. VideoScore: Building automatic metrics to simulate fine-grained human feedback for video generation. \emph{EMNLP}, pp.2105--2123, 2024. doi:10.18653/v1/2024.emnlp-main.127. \url{https://aclanthology.org/2024.emnlp-main.127/}.

\bibitem{videorewardbench} J. Liu, G. Liu, J. Liang et al. Improving video generation with human feedback. \emph{NeurIPS}, 2025. \url{https://arxiv.org/abs/2501.13918}.

\bibitem{videoscore2} X. He, D. Jiang, P. Nie et al. VideoScore2: Think before you score in generative video evaluation. \emph{arXiv:2509.22799}, 2025. \url{https://arxiv.org/abs/2509.22799}.

\bibitem{ffmpegfilters} FFmpeg developers. FFmpeg filters documentation, \texttt{freezedetect}. Accessed 24 October 2026. \url{https://ffmpeg.org/ffmpeg-filters.html\#freezedetect}.

\bibitem{ffmpegsource} FFmpeg developers. \texttt{vf\_freezedetect.c}, release \texttt{n7.0.2}. \url{https://github.com/FFmpeg/FFmpeg/blob/n7.0.2/libavfilter/vf_freezedetect.c}.

\bibitem{seeingfastslow} Y.-S. Wu et al. Seeing Fast and Slow: Learning the Flow of Time in Videos. \emph{ECCV}, 2026. Official source and released speed-change checkpoint: \url{https://github.com/Seeing-Fast-and-Slow/Seeing-Fast-and-Slow}.

\bibitem{finitekernel} T. Radzikowski and M. Modrzejewski. Finite-Sample Sensitivity of Kernel Metrics for Generative Evaluation. \emph{arXiv:2610.34477}, 2026. \url{https://arxiv.org/abs/2610.34477}.

\bibitem{evalcrafterrelease} Y. Liu, X. Cun, X. Liu et al. EvalCrafter: Benchmarking and evaluating large video generation models. \emph{CVPR}, pp. 22139--22149, 2024. Human annotations and videos: \url{https://github.com/EvalCrafter/EvalCrafter}.

\bibitem{i3d} J. Carreira and A. Zisserman. Quo vadis, action recognition? A new model and the Kinetics dataset. \emph{CVPR}, pp. 6299--6308, 2017.

\bibitem{videomaev2} L. Wang, B. Huang, Z. Zhao et al. VideoMAE V2: Scaling video masked autoencoders with dual masking. \emph{CVPR}, pp. 14549--14560, 2023.

\bibitem{vjepa} A. Bardes, Q. Garrido, J. Ponce et al. Revisiting feature prediction for learning visual representations from video. \emph{Transactions on Machine Learning Research}, 2024.

\bibitem{energy} G. J. Sz\'ekely and M. L. Rizzo. Energy statistics: A class of statistics based on distances. \emph{Journal of Statistical Planning and Inference} 143(8):1249--1272, 2013.

\bibitem{randomperm} J. Hemerik and J. Goeman. Exact testing with random permutations. \emph{TEST} 27:811--825, 2018.

\bibitem{dentonfergus} E. Denton and R. Fergus. Stochastic video generation with a learned prior. \emph{ICML}, PMLR 80:1174--1183, 2018. \url{https://proceedings.mlr.press/v80/denton18a.html}.

\bibitem{vbench2paper} D. Zheng, Z. Huang, H. Liu et al. VBench-2.0: Advancing video generation benchmark suite for intrinsic faithfulness. \emph{arXiv:2503.21755}, 2025.

\end{thebibliography}
}
\end{document}